# Audio Analytics-based Human Trafficking Detection Framework for Autonomous Vehicles

**Sagar Dasgupta**\*
Ph.D. Student
Department of Civil, Construction & Environmental Engineering
The University of Alabama
3014 Cyber Hall, Box 870205, 248 Kirkbride Lane, Tuscaloosa, AL 35487
Tel: (864) 624-6210; Email: sdasgupta@crimson.ua.edu

**Kazi Shakib**
Ph.D. Student
Department of Civil, Construction & Environmental Engineering
The University of Alabama
3014 Cyber Hall, Box 870205, 248 Kirkbride Lane, Tuscaloosa, AL 35487
Tel: (659) 210-0420; Email: khshakib@crimson.ua.edu

**Mizanur Rahman, Ph.D.**
Assistant Professor
Department of Civil, Construction & Environmental Engineering
The University of Alabama
3015 Cyber Hall, Box 870205, 248 Kirkbride Lane, Tuscaloosa, AL 35487
Tel: (205) 348-1717; Email: <u>mizan.rahman@ua.edu</u>

**Silvana V Croope**, Ph.D., ENV SP
Resilience Research Engineer
Alabama Transportation Institute
The University of Alabama
3051 Cyber Hall, Box 870205, 248 Kirkbride Lane, Tuscaloosa, AL 35487
Tel: (302)-593-5536; Email: svcroope@ua.edu

**Steven Jones, Ph.D.**
Director, Transportation Policy Research Center
Professor, Department of Civil, Construction & Environmental Engineering
The University of Alabama
3024 Cyber Hall, Box 870205, 248 Kirkbride Lane, Tuscaloosa, AL 35487
Tel: (205)-348-3137; Email: sjones@eng.ua.edu

\*Corresponding author

## ABSTRACT

Human trafficking is a universal problem, persistent despite numerous efforts to combat globally. Individuals of any age, race, ethnicity, sex, gender identity, sexual orientation, nationality, immigration status, cultural background, religion, socio-economic class, and education can be a victim of human trafficking. With the advancements in technology and the introduction of autonomous vehicles (AVs), human traffickers will adopt new ways to transport victims, which could accelerate the growth of organized human trafficking networks, whcih can make detection of trafficking in persons more challenging for law enforcement agencies. The objective of this study is to develop an innovative audio analytics-based human trafficking detection framework for autonomous vehicles. The primary contributions of this study are to: (i) define four non-trivial, feasible, and realistic human trafficking scenarios for AVs; (ii) create a new and comprehensive audio dataset related to human trafficking with five classes—i.e., crying, screaming, car door banging, car noise, and conversation; and (iii) develop a deep 1-D Convolution Neural Network (CNN) architecture for audio data classification related to human trafficking. We have also conducted a case study using the new audio dataset and evaluate the audio classification performance of the deep 1-D CNN. Our analyses reveal that the deep 1-D CNN can distinguish sound coming from a human trafficking victim from a non-human trafficking sound with an accuracy of 95%, which proves the efficacy of our framework.

**Keywords:** Human trafficking, Autonomous vehicle, Audio analytics.

## INTRODUCTION

Human trafficking is a global epidemic. People of any age, gender identities, and ethnicities from all across the world are constantly under threat of being victim of human trafficking. According to the Department of Homeland Security, falsification or threat of force is used to acquire cheap labor or commercial sex acts in human trafficking (*1*). In the latest reports of International Labor Organization, 40.3 million people are estimated to be forced in Human Trafficking in 2016 (*2*). COVID-19-related incidents have led to an increase in this heinous trend and the number of victims (*3*). With the advancements in technology and the introduction to autonomous vehicles (AVs), human traffickers will adopt new ways to transport the victims, which could accelerate the growth of organized human trafficking networks, and it can made detection of human trafficking more challenging for the law enforcement agencies. The primary challenge for detecting human trafficking utilizing AVs is that AVs may not be frequently checked by law enforcement agencies as AVs are expected to strictly follow traffic rules. Thus, human traffickers may use AVs to transport victims to avoid detection. AVs will fend off outside attacks, such as accidents and traffic-clogging; however, they may be vulnerable and thus manipulated by inside attackers like human traffickers (*4*). Thus, it is necessary to identify threats from within AVs.

Given the fact that ride-sharing AVs only have a limited amount of information related to their passengers, they can be exploited in human trafficking activities. In addition, as legislation related to AVs are underdeveloped, legislators also may find it difficult to hold AV manufacturers liable for such crimes because of limited ability for the Avs, and thus their manufacturers, to even be aware of human trafficking (*4*). Thus, in-vehicle surveillance to avoid human trafficking has become a topic of research. In general, it is a challenge to find any indication of human trafficking as human traffickers may threaten the victims to act sensibly outside vehicle. However, facial expression, voice tones, and monitoring stress of passengers inside an AV may help to identify a human trafficking event (*5*). Aggressive tones and behavioral patterns may indicate violence and illicit crimes like human trafficking. Identification of such behavioral patterns or tones may lead to finding large organizations of human traffickers and eventually revealing their routes and network.

An artificial intelligence-based computer vision strategy could be an effective approach for detecting the human trafficking event. However, an in-vehicle computer vision system for facial recognition has privacy issues as it stores facial data (*6*). In addition, vision-based models fail to recognize human traffickers with masks/veils or in foggy lighting condition, in blurry vision due to motion and low-resolution videos. On the other hand, with the advancement of audio analytics, detection of crying, screaming, or door banging sounds inside an AV can help to identify an event of human trafficking. For example, a victim may figure out that he/she is being trafficked in the middle or end of the trip using an AV and react in terms of crying, screaming, or door banging. However, there are several challenges the use of audio for the human trafficking scenarios: (i) there is no defined non-trivial, feasible, and realistic human traffic scenarios for AVs where audio analytics can be effective for human traffic event detection; (ii) there is no audio dataset related to human trafficking; and (iii) there is no appropriate architecture of deep learning model for audio data classification related to human trafficking.

The objective of this study is to develop an innovative audio analytics-based human trafficking detection framework for autonomous vehicles. The primary contributions of this study are to: (i) define four non-trivial, feasible and realistic human traffic scenarios for AVs; (ii) create a new and comprehensive audio dataset related to human trafficking with five classes—i.e., crying, screaming, car door banging, car noise, and conversation; and (iii) develop a deep 1-D Convolution Neural Network (CNN) architecture for audio data classification related to human trafficking. In this paper, we have conducted a case study using the new audio dataset and evaluate the audio classification performance of the deep 1-D CNN.

## RELATED WORK

Human trafficking can be detected in the victim recruitment phase or during the transportation of the victims. Law enforcement agencies and the department of transportation in every states are working together to detect and prevent human trafficking (*7*, *8*). Text mining, natural language processing, web scrapping, and statement analysis are proposed by researchers to detect human trafficking intentions in the recruitment phase (*6*, *9–15*). These methods can help to prevent human trafficking when it is still in a very early stage. However, it is difficult to prove human trafficking intention based on random internet advertisements or someone's chat. It is of utmost importance to intercept the human traffickers before they take control over the victim, more precisely saving the victims in the transportation phase. One of the techniques of identifying the human traffickers and victims is by real time facial recognition leveraging computer vision technology. Artificial intelligence algorithms, such as Convolution Neural Network (CNN) architecture (*16–18*) are widely used for facial recognition. The facial recognition requires the photo of human trafficker and the victim, which are not available in the database most of the time. On top of that if the trafficker wear masks, puts on veils or in the case uneven lighting condition(foggy), or blurry vision due to motion and low-resolution videos, facial recognition fails to detect human trafficker or the victim. Besides having facial information of individuals without sharing privacy policies creates massive privacy concerns (*19*, *20*) and even lose the autonomy over facial information.

One of the characteristics of a human trafficking event is the acoustic signature of the event. The victim's reaction to the trafficking event can result in crying, screaming or even trying to get rid of the vehicle. If such human trafficking event related sounds can be detected by analyzing the in-vehicle sounds, it can help detecting human trafficking incidents in real-time. Though sound or acoustic analysis for various purposes can be found in the literature, as far as the authors are concerned, study related to detecting human trafficking events with audio analytics has not been performed before. The existing sound identification and classification methods predominantly uses deep learning models and CNN is the most used algorithm and mel-frequency cepstral coefficient (MFCC) is preferred for feature extraction. In (*20*), a 1-D CNN with MFCC is used for detecting and classification illegal tree cutting related sound. In (*21*), a 1-D CNN with MFCC is used detect and classify illegal tree cutting related sound. A combination of 2D CNN and two Recurrent Neural Network (RNN) layers is used by (*22*) for obtaining better classification accuracy. Resource adaptive CNN (RACNN) model is used in (*23*) uses log mel-preprocessed data but significant improvement in terms of classification accuracy is not achieved. A 98.24% accuracy in siren sound detection is achieved by (*24*), by implementing a CNN based ensemble model called SirenNet combined with log mel and MFCC. A deep 1-D Convolution Neural Network (CNN) framework for audio data categorization related to human trafficking is introduced in this paper. We also performed a case study with the new audio dataset to assess the deep 1-D CNN's audio classification efficiency.

## HUMAN TRAFFICKING THREAT MODEL

Four non-trivial, feasible, and realistic human traffic scenarios are presented where an AV is used for transporting victims of human trafficking. In the scenarios, the victim gets into an AV to reach his/her desired destination. Note that the victim can be the only user of the AV or victim can share the trip with the trafficker without realizing the intention of the other AV users. The scenarios are based on two assumptions. The first assumption is traffickers can spoof AV GPS (Global Positioning System) receiver and redirect the AV to a different location instead of the user's (here victim) desired destination (*25*, *26*). The second assumption is that the victim figures out that he/she is being trafficked in the middle or end of the trip and react in terms of crying, screaming, or door banging. In scenario 1, the victim alone uses the AV and figures out the danger in the middle of the trip. In scenario 2, the victim is the only user of the AV, same as scenario 2, however, he/she recognize the danger at the end of the trip. In scenario 3, the victim shares the trip with another user who is part of the trafficker's gang without knowing and figures out the danger in the middle of the trip. Finally, in scenario 4, the victim and the trafficker use the same AV and the victim figures out the danger at the end of the trip. All these scenarios are discussed below.

### Scenario #1:
In the first scenario, the victim books an AV from origin 'O' to destination 'D'. The victim is a local of the city and he/she knows the roads well. After starting the trip, the victim relaxes and starts enjoying in vehicle entertainment options. At a point 'X', on the route between 'OD' the trafficker spoofs the AV GPS receiver and redirect the AV towards the trafficker's desired location "Y". After some time of the spoofing, the victim figures out that the AV is in a wrong route and figure out that he/she is in danger. As part of a reactive response the victim starts to cry and scream out of panic. He/she also tries to break out of the door. However, fails to get out of the car. The AV reaches location 'Y' and the traffickers force the victim to follow them to the next location of the trafficking network, taking the control over the victim for the purpose of exploitation.

### Scenario #2:
In the second scenario, the victim books an AV from 'O' to 'D' and at location 'X' the trafficker spoofs the AV's GPS receiver and redirect the AV towards location 'Y' same as scenario 1. But in scenario 2 the victim is not a local and have very little to no idea about the destination pathway (the road features). As a result, the victim fails to detect any change in the route and when the AV reaches location 'Y', he/she suddenly find himself/herself in a place where he/she did not expect to be. Then the victim starts crying and screaming out of panic. The traffickers get the victim out of the AV and force him/her to comply with them.

### Scenario #3:
In the third scenario, the victim uses a shared AV service where multiple users share a single AV such as the bus service these days. The victim gets into such an AV at 'O' to 'D'. Other passengers join the victim at different locations. The victim knows none of the other passengers. The other passengers are part of a trafficking gang, and they use fake identification (ID) to book the trip. At a certain point the victim starts feeling suspicious about one or multiple other passengers and starts feeling unsafe around them. The victim then the victim tries to drop off from the AV at the next stoppage other than the actual one. When the victim tries to do that, the gang members force  the victim to stay in the AV so that they can take the victim into the desired location 'Y'. As a result, the victim starts screaming, crying, and banging on the door.

**Scenario #4:**
In the fourth scenario, the victim uses a shared AV service same as scenario 3. When the AV reaches location 'Y', the gang members force the victim to get out of the AV and go with them. As a result, the victim panics and stars crying and screaming. The traffickers gain control over the victim and take him/her to the next location of the trafficking network.

## DATASET GENERATION

The presented deep learning-based framework for detecting human trafficking is trained and evaluated using the Human Trafficking Audio Dataset (HTA Dataset). We have created a new HTA dataset for the first time as there is no HTA dataset available currently as per the authors knowledge.  The HTA dataset is intended to address the lack of human trafficking related audio dataset. During the creation of the dataset, the audio classes are chosen based on the premise that audio from both regular and HTA environments should be included to make the dataset comprehensive. Special importance is given to the detail of the audio samples so that they mimic real-life human trafficking events.

The HTA dataset is consists of five classes: crying, screaming, car door banging, car noise experienced form the inside of the car and car passenger conversations (see **Figure 1**). The first three classes are related to HT events, and the last two classes related to non-HT event. Crying and screaming are common reactions from a human trafficking victim after the victim realize that he/she is under illegal control of human trafficker or abuser. Both the crying and screaming classes include audio from victims of all age and gender groups (i.e., audio of adult, teen, and children including both male and female). After recognizing the danger, a victim can try to get out the vehicle and try to bang on the door, hence car door banging from inside an AV is also included as a class. In order to distinguish between a HT event and a non-HT event, two classes related to non-HT are also included. A vehicle noise class includes audio of left/right turn indicators, vehicle starting, and engine idling. Conversations between passengers of all age group and gender are included in the conversation class and the conversations are recorded from the inside of the vehicle. To create the dataset, related YouTube audio files are used. The audio files are truncated to multiple files of 2-3 seconds audio. Examples of raw audio files of each class are shown in **Figure 2**, where x-axis represents audio sample number and y-axis represents the frequency. The HTA dataset includes audio files and metadata. The metadata includes, the audio file name, class identification number and the class details.

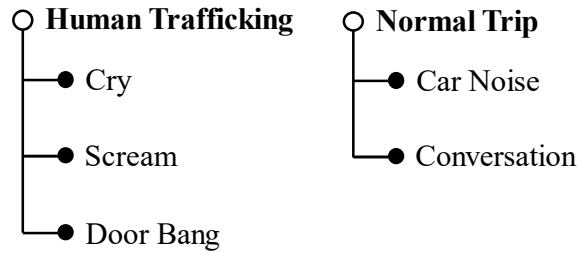

**Figure 1 Layers of HTA audio dataset**

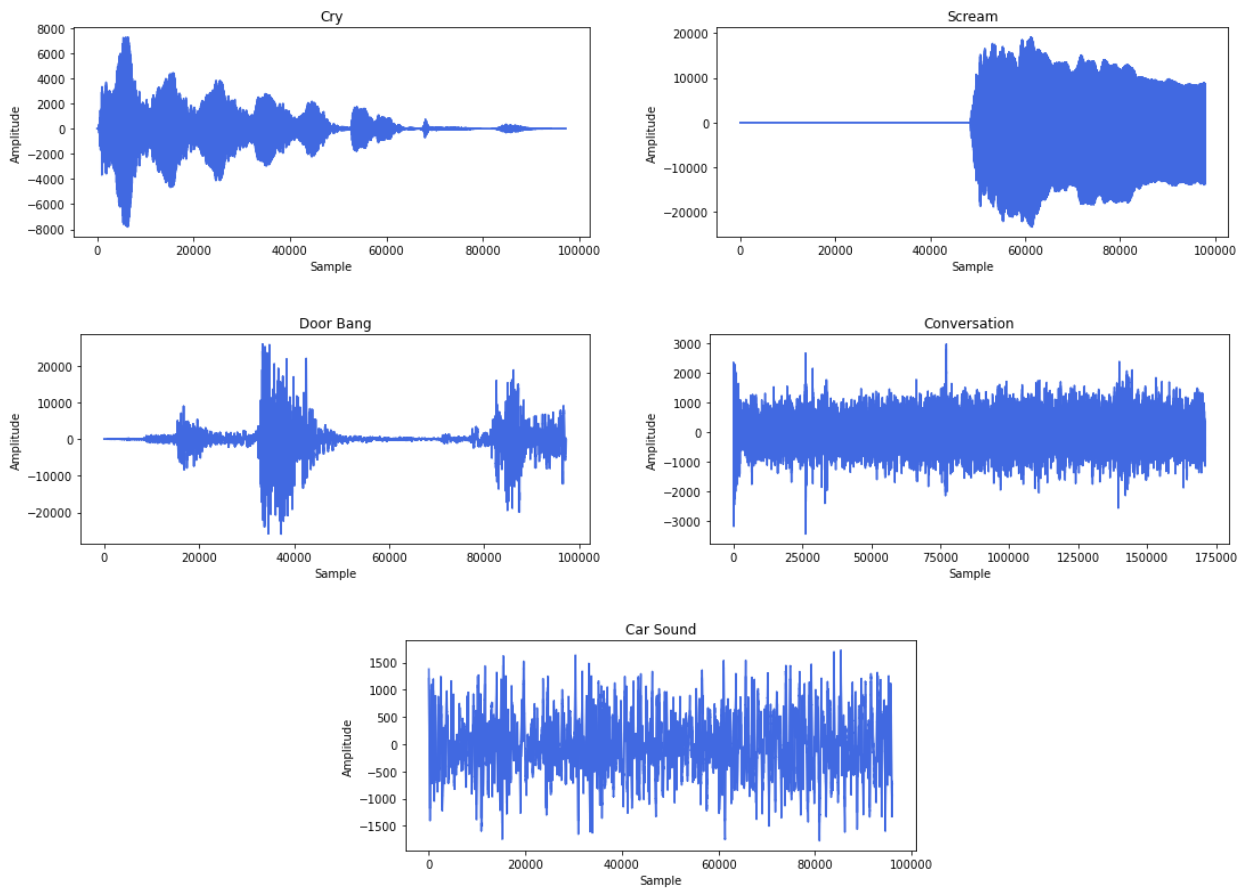

**Figure 2 Raw audio samples of five classes of HTA dataset**

## DETECTION FRAMEWORK

**Figure 3** presents a human trafficking detection framework that uses deep learning classification model to classify audio data to detect human trafficking incidents. We assume that an in-vehicle microphone will be used to record sounds related to a human trafficking event. After that, the recorded soundwave is preprocessed, and features are extracted for classification. The extracted features are then fed into a deep Convolutional Neural Network (CNN) model. The CNN model is trained with different classes of sound that includes sounds related to incidents of human trafficking using AVs as well as sounds related to normal AV trips. The CNN model is used to classify types of sounds and if the type matches with human trafficking related sound, then an alert message is generated; otherwise, no action is taken. The audio analytics module is connected to an app which continuously track the location of the vehicle. When a human trafficking incident is detected by the audio analytics module, the app will continuously transit alert messages and AVs real time location information to the law enforcement agency. As soon as the alert is received by the law enforcement agencies, they act instantaneously and interdict the trafficker.

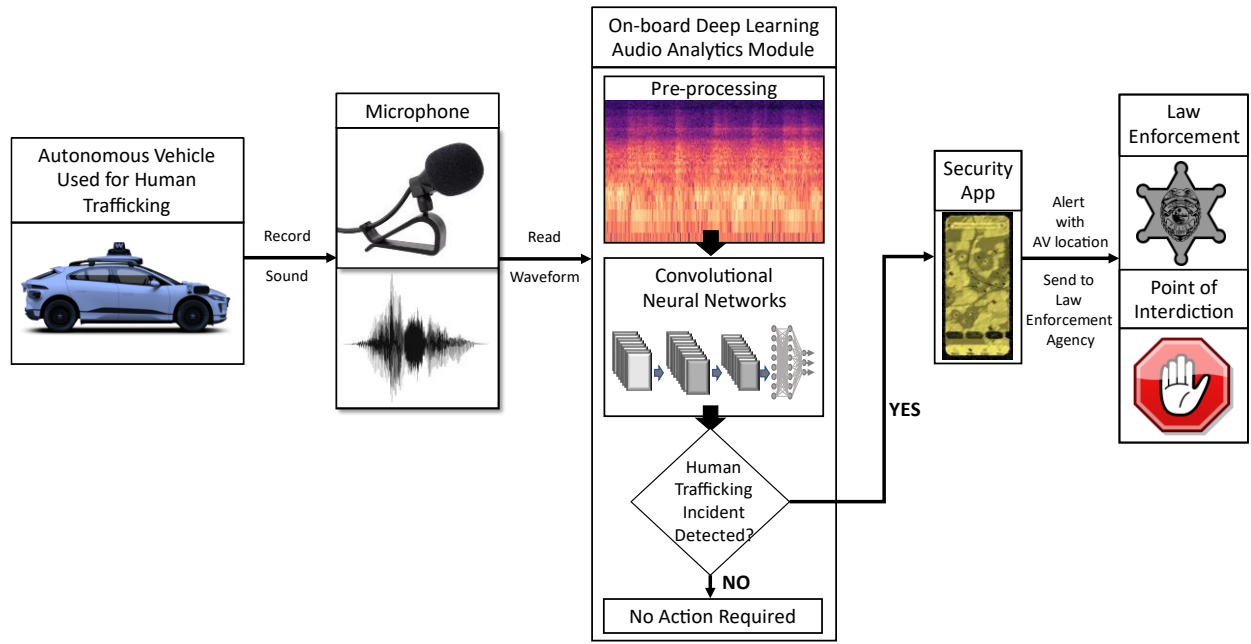

**Figure 3 Human trafficking detection framework**

The audio classification architecture is shown in **Figure 4**. The HTA dataset consists of audio files of five classes, and a metadata file containing the audio file name and corresponding class information. The audio files are of 2-3 seconds in length. Audio files are split into training and validation data. 80% audio files are used for training and 20% files are used for validation. Both training and validation dataset contains samples from all the classes, and they are balanced. The splitting is performed in the metadata file. All the audio samples are pre-processed before they are fed to the CNN classification model. The files are processed and used for training as batches to optimize the use of computer hardware memory. The first step in the audio sample preprocessing is to convert all the audio files into stereo format that ensures that all the audio sample files are of same dimension. Then the samples are resampled at a rate of 44100Hz. After resampling, samples are padded with silence to make them of same length. The samples are then augmented by performing a time shift of a factor of 0.4.

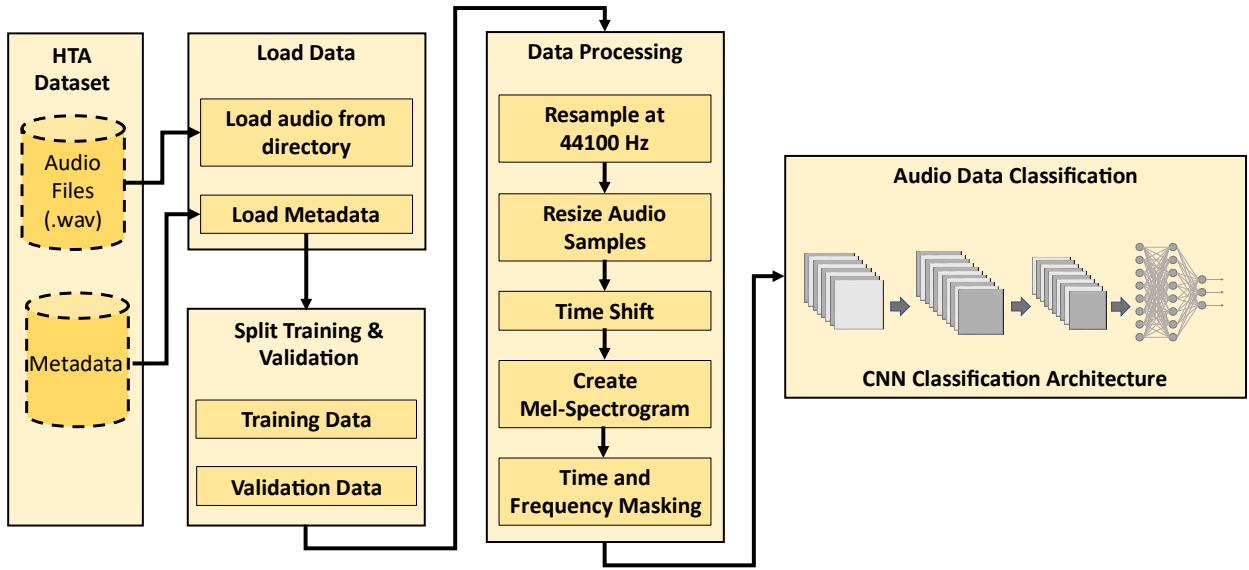

**Figure 4 Audio classification architecture**

After pre-processing the samples, features need to be extracted. CNN is an effective model to extract features from images and then further classify or recognize the image. As a result, it is necessary to convert the audio samples to image. The mel-spectrogram can extract the hidden features from an audio file and recreate it as an image. The mel-spectrogram converts the audio file to a linear audio spectrogram by mapping hertz values to the mel scale that is a logarithmic transformation of a signal frequency. Hence, audio sample is converted to mel-spectrogram (see **Figure 5**) using Librosa (*27*), a python package. The MFCC have smaller dimension than the mel-spectrogram that extracts the most important features related to the audio sample. Then spectrogram augmentation is performed. In spectrogram augmentation, a range of frequencies are masked which adds horizontal bars on the spectrograms. Moreover, ranges of time are masked adding vertical bars in the spectrogram. After data augmentation, the data is normalized and fed to the CNN classification model.

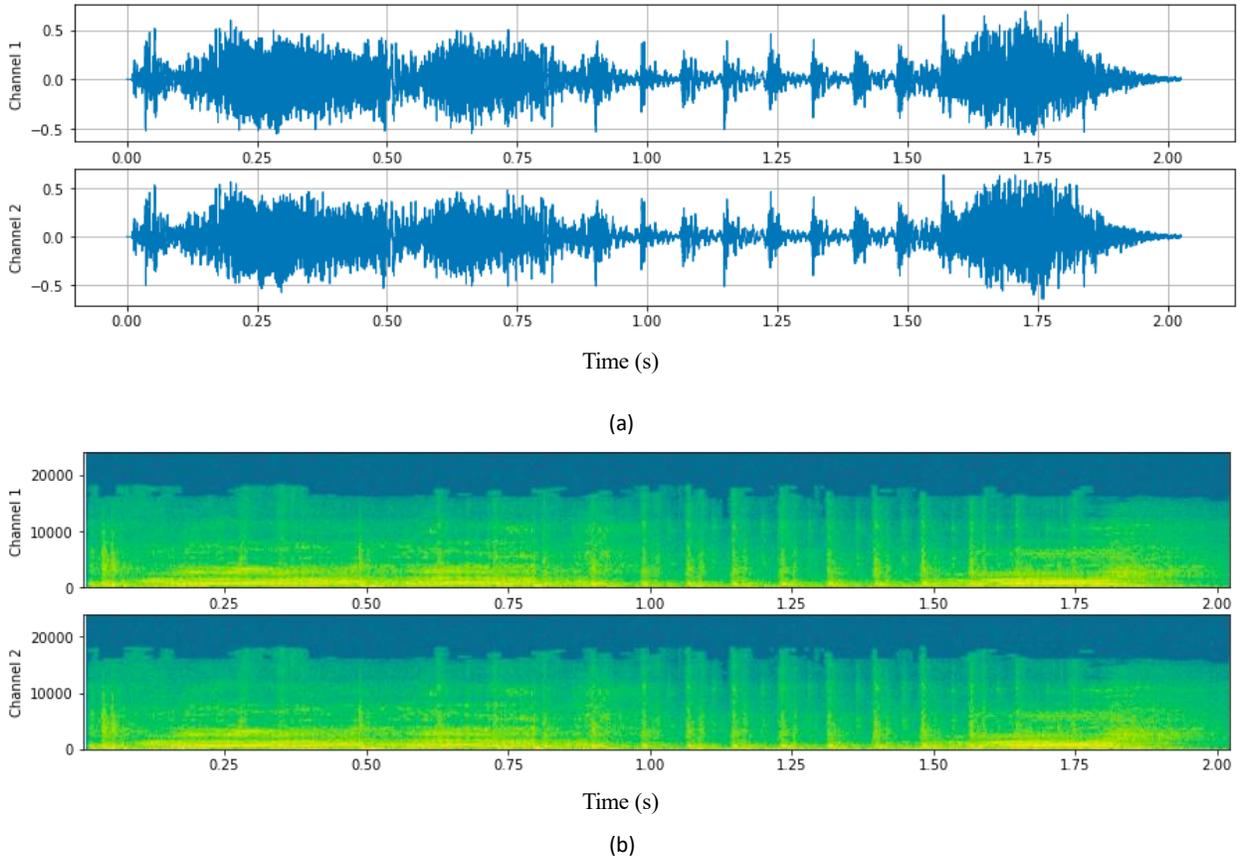

**Figure 5. Sample raw audio data (a) and (b) corresponding spectrogram**

## AUDIO ANALYTICS MODEL DEVELOPMENT AND EVALUATION RESULTS

**Figure 6** represents the deep 1-D CNN classification architecture. The presented model contains more than 1 million parameters. It consists of four convolution (CNN) layers of four different shapes. The filter size of the CNN layers is 32, 64, 128 and 256 respectively. The filters create border effect while reducing the number of feature or pixels. To deal with the border effect, padding and stride are implemented. Padding adds pixels to the edge of the image. The padding is (2,2) in all four CNN layers. The kernel size is (3,5) for first two CNN layers and (5,5) for next two CNN layers. The stride is (2,2) for first CNN layer and (1,1) for rest of the CNN layers. Stride (2,2) corresponds to movement of the filter two pixels right for each horizontal movement and two pixels down for each vertical movement. In the CNN layers *tanh* is used as the activation function that ensures strong gradients during training and faster learning. Stochastic Gradient Descent (SGD) algorithm is used as the optimizer. In SGD weights are updated for each training sample and it also reduces the computational weight. 2D batch normalization is also performed in each CNN layer to reduce the training time. The above mentioned hyperparameters are presented in Table 1 and selected by trial-and-error method where the higher model testing accuracy is the deciding factor. After the CNN layers, the data is pooled using 2D average pooling function to down sample the data by replacing the features by average values over input window. Then, the audio data is classified by a single connected linear layer and the output layer has five features, same number as the number of input classes. **Figure 7** shows the loss function and accuracy value over the epochs. The x-axis represents number of epochs, the primary y-axis represents the loss and the secondary y-axis represents the accuracy. Cross entropy loss in pytorch is used for calculating the loss. With increasing epochs, the loss decreases and stabilized near 100 epochs and training accuracy increases with number of epochs.

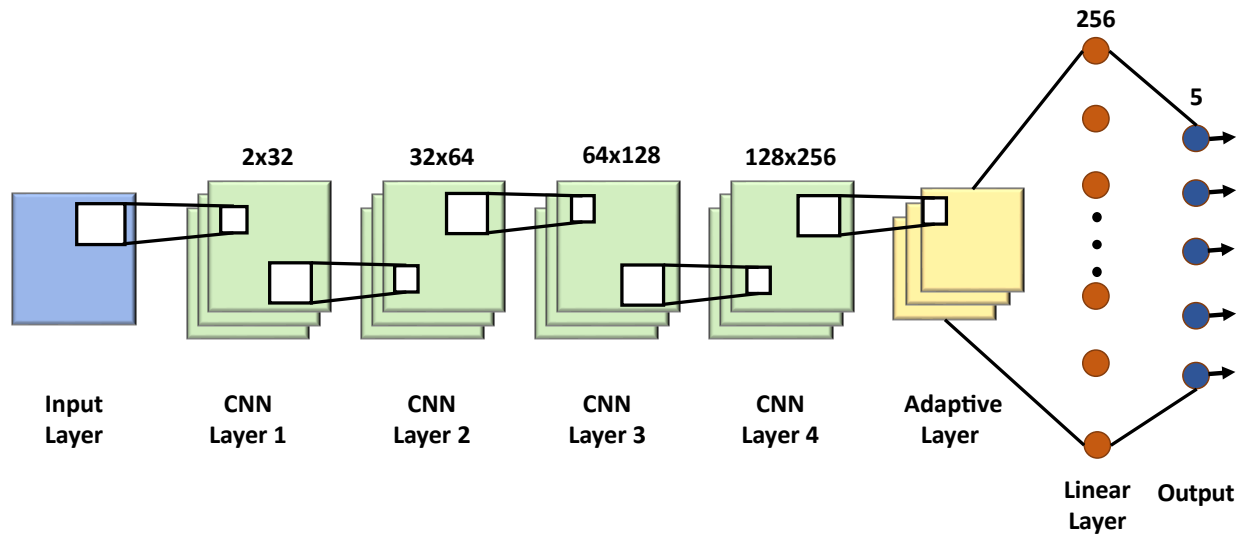

**Figure 6 CNN classification architecture**

**TABLE 1. CNN Model Hyperparameters**

| Hypermeters and Optimizer | Value |
|---|---|
| CNN layer 1 | 2 x 32 |
| CNN layer 2 | 32 x 64 |
| CNN layer 3 | 64 x 128 |
| CNN layer 4 | 128 x 256 |
| Adaptive layer | 2D Average Pooling |
| Linear layer (Classification) | 256 |
| Number of epochs | 100 |
| Batch size | 16 |
| Activation function | tanh |
| Optimizer | SGD (stochastic gradient descent) |

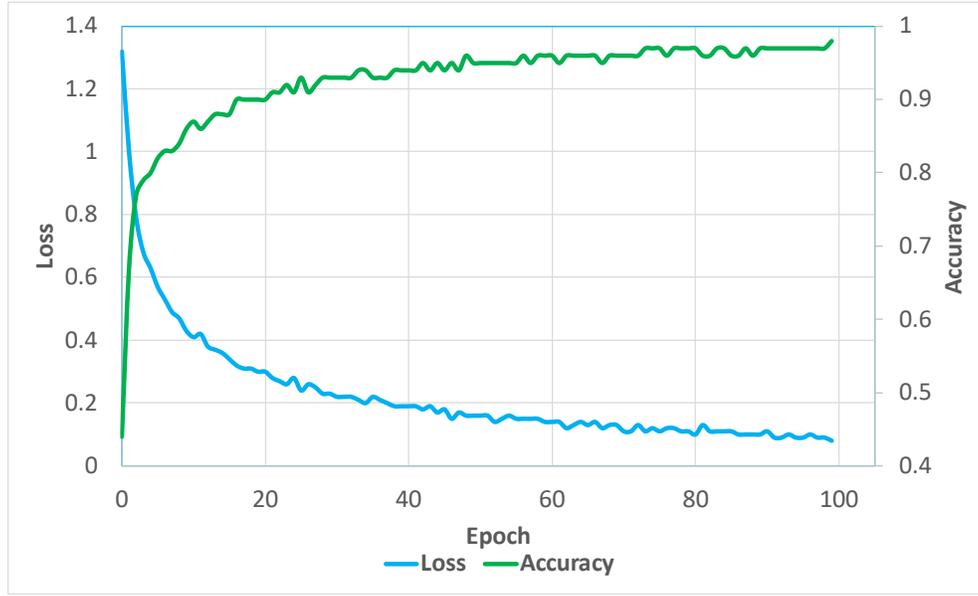

**Figure 7 CNN Training Performance**

The presented model is trained for 100 epochs and the training and testing accuracy are 97% and 94.98%. The training accuracy profile is shown in **Figure 7**. The confusion matrix that shows the model performance for all five classes is presented in **Figure 8**. The x-axis presents the predicted label (i.e., predicted class) and the y-axis presents the true label (i.e., true class). Based on the confusion matrix our model can detect car sound (i.e., internal car noise) with 100% accuracy which indicates that vehicle's sound (car indicators sound, car starting sound, engine idling sound) will not be detected as a human trafficking event. Another audio class, that doesn't indicate a human trafficking event is conversation of the passengers. Our model can detect conversations with 97% accuracy. There is a possibility of false detection as conversation can be detected as door bang and cry. However, the possibility is significantly low as there is only 1% chance of detecting conversation sound as door banging or crying.

The three audio classes that will lead to the detection of a human trafficking are cry, scream and door bang. Based on the model testing performance, the door banging sound is detected with 93% accuracy. The model confuses between door banging and scream sound for rest 7%. Though the accuracy is not 100%; however, both door banging and screaming sounds are detected as a human trafficking incident. Thus, there will be no false alarm. The model can detect cry sound with 93% accuracy and again the model classifies cry sound as scream sound for 7% samples. However, as both cry and scream classes detect a human trafficking incident, our model detects the event as a human trafficking event and send alert to the law enforcing agencies. The classification accuracy for screaming sound is 89% , which is lowest among all the classes. For 1% samples, scream audio is falsely detected as conversation which may lead to false alarm. For the remaining 10% of samples, however, scream sounds are detected as cry class. As both scream and cry classes are used to detect human trafficking event, there will be no false alarm.

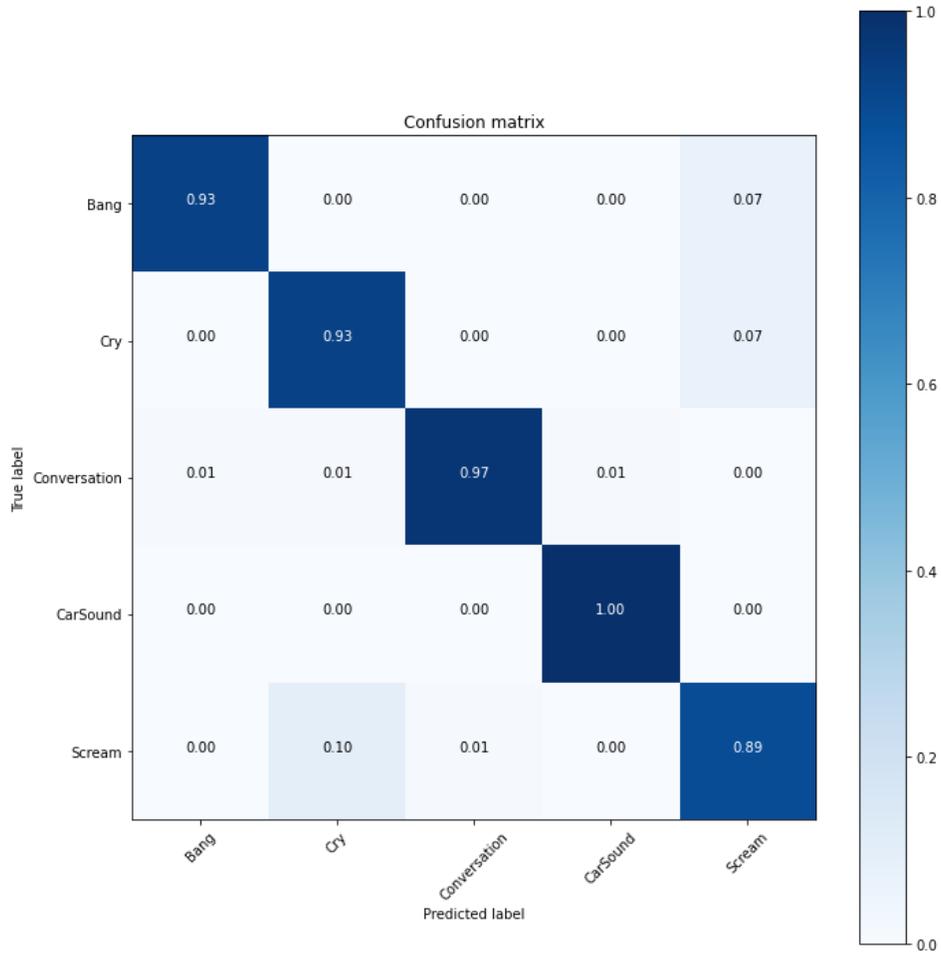

**Figure 8. Confusion matrix**

## CONCLUSION

In this study, we have presented an innovative human trafficking detection framework using audio analytics for AVs. However, there are several challenges the use of audio for the human trafficking scenarios related to defining non-trivial, feasible and realistic human traffic scenarios for AVs where audio analytics can be effective for human traffic event detection, creating audio dataset related to human trafficking, and developing appropriate architecture using deep learning model for audio data classification related to human trafficking with high accuracy. We have defined four non-trivial, feasible and realistic human traffic scenarios for AVs using audio analytics. We have also created a new and comprehensive HTA dataset related to human trafficking with five classes (i.e., crying, screaming, car door banging, car noise, and conversation). After that, we have developed a deep 1-D CNN architecture for audio data classification related to human trafficking. Finally, we have conducted a case study using the HTA dataset and evaluated the audio classification performance of the deep 1-D CNN. Our analyses reveal that the deep 1-D CNN can distinguish from a human trafficking related sound from a non-human trafficking sound with an accuracy of 95%, which proves the efficacy of our framework. Although there are some instances of wrong classification, the overall human trafficking incident detection accuracy is high with a low false negative. In terms of future research, the dataset can be enriched with more diverse classes and the model can be further optimized for higher accuracy.

## ACKNOWLEDGMENTS

This material is based on a study supported by the Alabama Transportation Institute (ATI). Any opinions, findings, and conclusions or recommendations expressed in this material are those of the author(s) and do not necessarily reflect the views of the Alabama Transportation Institute (ATI).

## AUTHOR CONTRIBUTIONS

The authors confirm contribution to the paper as follows: study conception and design: S. Dasgupta, M. Rahman, S. Croope; data collection: S. Dasgupta, K. Shakib; interpretation of results: S. Dasgupta, M. Rahman, and S. Jones; draft manuscript preparation: S. Dasgupta, K. Shakib, M. Rahman, S. Croope, and S. Jones. All authors reviewed the results and approved the final version of the manuscript.